\pdfoutput=1
\documentclass{article}

    \usepackage[final,nonatbib]{neurips_2022}

\usepackage{amsmath,amsfonts,bm}

\def\eqref#1{equation~\ref{#1}}

\def\floor#1{\lfloor #1 \rfloor}
\def\1{\bm{1}}

\def\vb{{\bm{b}}}

\def\vh{{\bm{h}}}

\def\vx{{\bm{x}}}

\def\mA{{\bm{A}}}

\def\mH{{\bm{H}}}

\def\mX{{\bm{X}}}

\DeclareMathAlphabet{\mathsfit}{\encodingdefault}{\sfdefault}{m}{sl}
\SetMathAlphabet{\mathsfit}{bold}{\encodingdefault}{\sfdefault}{bx}{n}

\def\gG{{\mathcal{G}}}

\newcommand{\E}{\mathbb{E}}

\newcommand{\R}{\mathbb{R}}

\DeclareMathOperator*{\argmax}{arg\,max}

\def\V{{\mathcal{V}}}
\def\E{{\mathcal{E}}}

\usepackage[utf8]{inputenc} 
\usepackage[T1]{fontenc}    
\usepackage{hyperref}       
\usepackage{url}            
\usepackage{booktabs}       
\usepackage{amsfonts}       
\usepackage{nicefrac}       
\usepackage{microtype}      
\usepackage{xcolor}         
\usepackage{color}
\usepackage{cases}

\usepackage{multirow}
\usepackage{microtype}
\usepackage{booktabs}
\usepackage[utf8]{inputenc}
\usepackage{listings}
\usepackage{caption}
\usepackage{dsfont}
\usepackage{graphicx}
\usepackage{algorithm}
\usepackage{algorithmic}
\usepackage{subfig}
\usepackage{ulem}
\usepackage{afterpage}
\usepackage{xspace}
\usepackage{adjustbox}
\usepackage{paralist}
\usepackage{fancybox}

\usepackage{amsmath}
\usepackage{amssymb}
\usepackage{mathtools}
\usepackage{amsthm}

\theoremstyle{plain}
\newtheorem{theorem}{Theorem}[section]

\newtheorem{lemma}[theorem]{Lemma}

\theoremstyle{definition}
\newtheorem{definition}[theorem]{Definition}

\newtheorem{property}[theorem]{Property}
\theoremstyle{remark}

\usepackage[textsize=tiny]{todonotes}

\newcommand{\aggr}{\textsc{AGGR}}

\newcommand{\sst}{\texttt{GraphSST2\xspace}}
\newcommand{\twitter}{\texttt{Twitter\xspace}}
\newcommand{\mutag}{\texttt{MUTAG\xspace}}
\newcommand{\bbbp}{\texttt{BBBP\xspace}}
\newcommand{\bace}{\texttt{BACE\xspace}}

\newcommand{\bamotif}{\texttt{BA2Motifs\xspace}}

\newcommand{\eval}{\textsc{Eval}\xspace}
\newcommand{\score}{\textsc{Score}\xspace}
\newcommand{\fidelity}{\textsf{\small Fidelity}\xspace}
\newcommand{\invfidelity}{\textsf{\small Inv-Fidelity}\xspace}
\newcommand{\sparsity}{\textsf{\small Sparsity}\xspace}
\newcommand{\normfidelity}{\textsf{\small N-Fidelity}\xspace}
\newcommand{\harfidelity}{\textsf{\small H-Fidelity}\xspace}
\newcommand{\norminvfidelity}{\textsf{\small N-Inv-Fidelity}\xspace}

\newcommand{\computehn}{\textsf{\small Compute-HN}\xspace}
\newcommand{\computehnmc}{\textsf{\small Compute-HN-MC}\xspace}

\newcommand{\method}{\textmd{GStarX\xspace}}

\title{GStarX: Explaining Graph Neural Networks with Structure-Aware Cooperative Games}

\author{Shichang Zhang$^1$
 \quad Yozen Liu$^2$ \quad Neil Shah$^2$ \quad Yizhou Sun$^1$\\ 
$^1$University of California, Los Angeles \quad
$^2$Snap Inc. \\
$^1$\texttt{\{shichang, yzsun\}@cs.ucla.edu} \quad
$^2$\texttt{\{yliu2, nshah\}@snap.com}}

\begin{document}

\maketitle
\begin{abstract}
Explaining machine learning models is an important and increasingly popular area of research interest.  The Shapley value from game theory has been proposed as a prime approach to compute feature importance towards model predictions on images, text, tabular data, and recently graph neural networks (GNNs) on graphs. In this work, we revisit the appropriateness of the Shapley value for GNN explanation, where the task is to identify the most important subgraph and constituent nodes for GNN predictions. We claim that the Shapley value is a non-ideal choice for graph data because it is by definition not structure-aware. We propose a \underline{G}raph \underline{St}ructure-\underline{a}wa\underline{r}e e\underline{X}planation (\method) method to leverage the critical graph structure information to improve the explanation. Specifically, we define a scoring function based on a new structure-aware value from cooperative game theory proposed by Hamiache and Navarro (HN). When used to score node importance, the HN value utilizes graph structures to attribute cooperation surplus between neighbor nodes, resembling message passing in GNNs, so that node importance scores reflect not only the node feature importance, but also the node structural roles. We demonstrate that \method{} produces qualitatively more intuitive explanations, and quantitatively improves explanation fidelity over strong baselines on chemical graph property prediction and text graph sentiment classification.\footnote{Code available at \color{cyan}{ \url{https://github.com/ShichangZh/GStarX}}} 

\end{abstract}

\section{Introduction}\label{sec:introduction}
Explainability is crucial for complex machine learning (ML) models in sensitive applications, helping establish user trust and providing insights for potential model improvements. Many efforts focus on explaining models on images, text, and tabular data. In contrast, the explainability of models on graph data is yet underexplored. Since explainability can be especially critical for many graph tasks like drug discovery, and interest in deep graph models is growing rapidly, further investigation of graph explainability is warranted. In this work, we study graph ML explanation with graph neural networks (GNNs) as the target models, given their popularity and widespread use for graph machine learning tasks \cite{ying2018graph, sankar2021graph, wu2020graph, tang2020knowing, tang2022friend, zhao2021synergistic}.

In ML explainability, important features are identified, and the Shapley value \cite{shapley} has been deemed as a ``fair'' scoring function for computing feature importance. Originally from cooperative game theory, many values, including the Shapley value, have been proposed for allocating a total payoff to players in a game. When used for scoring the feature importance of a data instance, the model prediction is treated as the total payoff and the features are considered as players. In particular, for an instance with $n$ features $\{\vx_1, \dots \vx_n \}$, the Shapley value of its $i$th feature $\vx_i$ is computed via aggregating $m(i, S)$, which are the marginal contributions of $\vx_i$ to sets of other features $\vx_S\subseteq {\{\vx_1, \dots, \vx_n\} \setminus \{\vx_i\}}$. Each $\vx_S$ is called a \textit{coalition}. Each $m(i, S)$ is computed as the difference between model outputs for $\vx_S \cup \{\vx_i\}$ and $\vx_S$, e.g., difference of probability belonging to a target class for these two set of features, and it is meant to capture the interaction between $\vx_i$ and $\vx_S$. The Shapley value is widely used for explaining ML models on images, text, and tabular data, when the features are pixels, words, and attributes \cite{shapley_regression, SHAP}.

The Shapley value has recently been extended to explain GNNs on graphs through feature importance scoring as above, where features are nodes \cite{graphsvx} or supernodes \cite{subgraphx}. We argue that the Shapley value is a non-ideal choice for (super)node importance scoring because its contribution aggregation is \textbf{non-structure-aware}. The Shapley value aggregation assumes no structural relationship between $\vx_i$ and $\vx_S$ even though they are both parts of the input graph (a review of the Shapley value is in Section \ref{subsec:cooperative_game}). Since the graph structure generally contains critical information and is crucial to the success of GNNs, we consider properly leveraging the structure with a better \textbf{structure-aware} scoring function.

\begin{figure*}[t]
\begin{center}
\includegraphics[width=\textwidth]{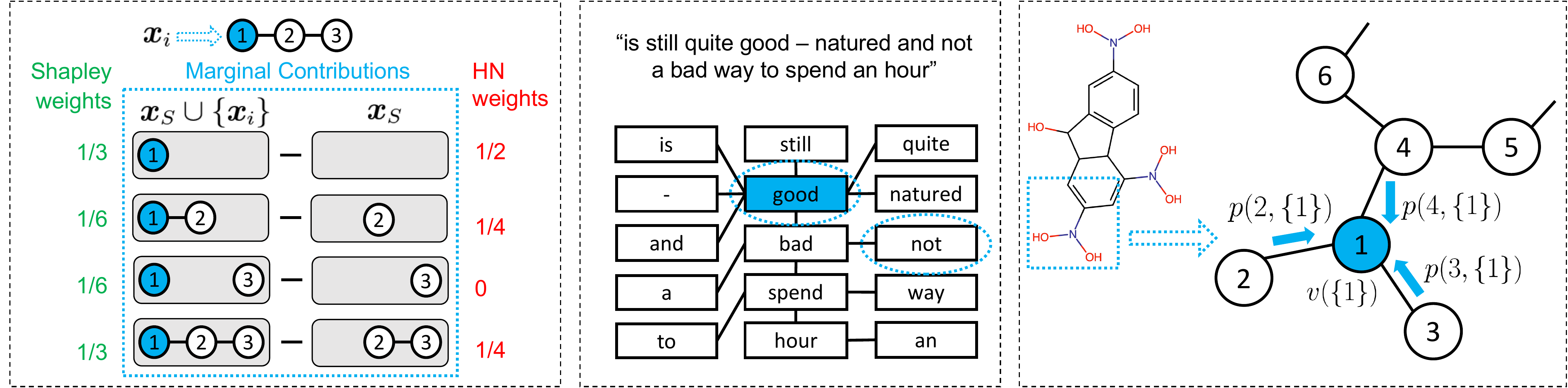}
\end{center}
\caption{Explanations on graphs with structure-aware values (like HN) offers advantages over non-structure-aware values (like Shapley). \textbf{(a) Synthetic graph (left):} The Shapley value assigns weights to $m(i, S)$ only based on size of $\vx_S$, while the HN value assigns weights considering structures and in particular gives zero weight to the disconnected $\vx_S$. \textbf{(b) Text graph (middle):} For a sentence classified as positive, the \{"not", "good"\} coalition shouldn't be considered when they are not connected by "bad". \textbf{(c) Chemical graph (right):} For a chemical graph with mutagenic functional group -NO2, the importance of the atom N (node 1) is better recognized if decided locally within the functional group. }
\vskip -0.2in
\label{figure:motivation_examples}
\end{figure*}

We propose \underline{G}raph \underline{St}ructure-\underline{a}wa\underline{r}e e\underline{X}planation (\method), where we construct a structure-aware node importance scoring function based on the Hamiache-Navarro (HN) value \cite{hn_value} from cooperative game theory. Recall that GNNs make predictions via message passing, during which node representations are learned by aggregating messages from neighbors. Message passing aggregates both feature and structure information, resulting in powerful structure-aware models \cite{gnn_structure_count}. The HN value shares a similar idea to message passing by allocating the payoff surplus generated from the cooperation between neighboring players (nodes). When used as a scoring function to explain node importance, the HN value captures both features and structural interactions between nodes (details in Section \ref{sec:methodology}). Figure \ref{figure:motivation_examples}\textbf{(a)} shows an example comparing the Shapley value and the HN value. In this example, their difference boils down to different aggregation weights of marginal contributions, where the former is uniform and the latter is structure-aware (details in Section \ref{subsec:motivation}). In summary, our contributions are:

\begin{compactitem}[\textbullet]
    \item Identify the non-structure-aware limitation of the Shapley value for GNN explanation.
    \item Introduce the structure-aware HN value from cooperative game theory to the graph machine learning community and connect it to the GNN message passing and GNN explanation.
    \item Propose a new HN-value-based GNN explanation method \method{}, and demonstrate the superiority of \method{} over strong baselines for explaining GNNs on chemical and text graphs.
\end{compactitem}

\section{Preliminaries} \label{sec:preliminary}
\subsection{Graph neural networks} 
Consider a graph $\gG$ with (feature-enriched) nodes $\V$ and edges $\E$. We denote $\gG$ as $\gG = (\V, \mX, \mA)$, where $\mX \in \R^{n \times d}$ denotes $d$-dimensional features of $n$ nodes in $\V$, and $\mA \in \{0,1\}^{n \times n}$ denotes the adjacency matrix specifying edges in $\E$. GNNs make predictions on $\gG$ by learning representations via the \textit{message-passing} mechanism. During message passing, the representation of each node $u \in \V$ is updated by aggregating its own representation and representations (messages) from its neighbors. We denote the set of neighbors as $\mathcal N(u)$. This aggregation is recursively applied, so $u$ can collect messages from its multi-hop neighbors and produce structure-aware representations \cite{gnn_structure_count}. With $\vh_i^{(l)}$ denotes the representation of node $i$ at iteration $l$, and $\aggr(\cdot, \cdot)$ denotes the aggregation operation, e.g. summation, the representation update is shown in Equation  \ref{eq:message_passing}.



\begin{equation} \label{eq:message_passing}
\vh_{u}^{(l)} = \aggr (\vh_{u}^{(l-1)}, \{\vh_i^{(l-1)} | i \in \mathcal N(u) \})
\end{equation} 

\subsection{Cooperative games} \label{subsec:cooperative_game}
\noindent\textbf{A cooperative game} denoted by $(N,v)$, is defined by a set of players $N = \{1, \dots, n\}$, and a \textit{characteristic function} $v: 2^{N} \rightarrow \R$. $v$ takes a subset of players $S \subseteq N$, called a \textit{coalition}, and maps it to a payoff $v(S)$, where $v(\emptyset) \coloneqq 0$. A \textit{solution function} $\phi$ is a function maps each given game $(N, v)$ to $\phi(N, v) \in \R^n$. The vector $\phi(N, v)$, called a \textit{solution}, represents a certain allocation of the total payoff $v(N)$ generated by all players to each individual, with the $i$th coordinate $\phi_i(N, v)$ being the payoff attributed to player $i$. $\phi(N, v)$ is also called the ``value'' of the game when it satisfies certain properties, and different values were proposed to name solutions with different properties \cite{shapley, core_econ}.  
 
\noindent\textbf{The Shapley value} is one popular solution of cooperative games. The main idea is to assign each player a ``fair'' share of the total payoff by considering all possible player interactions. For example, when player $i$ cooperates with a coalition $S$, the total payoff $v(S \cup \{ i \})$ may be very different from $v(S) + v(\{ i \})$ because of $i$'s interaction with S. Thus the marginal contribution of $i$ to $S$ is defined as by $m(i, S) = v(S \cup \{ i \}) - v(S)$. Then the formula of the Shapley value for $i$ is shown in Equation \ref{eq:shapley}, where marginal contributions to all possible coalitions $S \subseteq N \backslash \{i\}$ are aggregated. The first identify in Equation \ref{eq:shapley} shows that the aggregation weights are first uniformly distributed among coalition sizes $k$ (outer average), then uniformly distributed among all coalitions with the same size (inner average).
\begin{align}  
    \phi_i(N, v) &= \overbrace{\frac{1}{n}\sum_{k = 0}^{n - 1}}^\text{Average over $k$} \overbrace{
    \frac{1}{\binom{n - 1}{k}} \sum_{\substack{S \subseteq N \backslash \{ i\} \\ |S| = k}}}^\text{Average over $S$ s.t. $|S| = k$} m(i, S) 
    &= \sum_{S \subseteq N \backslash \{ i\}}
    \frac{|S|!(n - |S| - 1)!}{n!} m(i, S)  \label{eq:shapley}
\end{align}

\noindent\textbf{Games with communication structures.} Although the Shapley value is widely used for cooperative games, its assumption of fully flexible cooperation among all players may not be achievable. Some coalitions may be preferred over others and some may even be impossible due to limited communication among players. Thus, \cite{myerson} uses a graph $\gG$ as the \textit{communication structure} of players to represent cooperation preference. A game with a communication structure is defined by a triple $(N, v, \gG)$, with $N$ being the node set of $\gG$. This game formulation is more practical than fully flexible cooperation when cooperation preference is available. Several values with different properties have been proposed for such games \cite{myerson, position_value, hamiache_value, d_myerson} including the HN value \cite{hn_value}.

\section{GNN explanation via feature importance scoring} \label{sec:motivation}
\subsection{Problem formalization} \label{subsec:feature_importance}
A general approach to formalize an ML explanation problem is through feature importance scoring \cite{SHAP, general-scoring1}, where features may refer to pixels of images, words of text, or nodes/edges/subgraphs of graphs. Let $f (\cdot)$ denote a to-be-explained GNN, $\gG = (\V, \mX, \mA)$ denote an input graph, and $0 < \gamma < 1$ denote a sparsity constraint to enforce concise explanation. GNN explanation via subgraph scoring is aimed to find a subgraph $g$ that maximizes a given evaluation metric \eval$(\cdot, \cdot, \cdot)$, which measures the faithfulness of $g$ to $\gG$ regarding making predictions with $f (\cdot)$, i.e.
\begin{equation} \label{eq:obj}
    g^* = \argmax_{g \subseteq \gG, |g| \leq \gamma |\gG|} \eval(f (\cdot), \gG, g)
\end{equation}
When the task is graph classification and $f (\cdot)$ outputs a one-sum vector $f(\gG) \in [0, 1]^{C}$ containing probabilities for $\gG$ belongs to $C$ classes, an example \eval can be the prediction probability drop for removing $g$ from $\gG$, i.e.  $\eval(f (\cdot), \gG, g) = \left[ f(\gG) \right]_{c^*} - \left[ f(\gG \backslash g) \right]_{c^*}$ with $c^*=\argmax_{c} [f(\gG)]_c$. 

In practice, since the number of subgraphs is combinatorial in the number of nodes, the objective is often relaxed to finding a set of important nodes or edges first and then inducing the subgraph \cite{gnnexplainer, pgexplainer, graphsvx}. A more tractable objective of finding the optimal set of nodes $S^* \subseteq \V$ \footnote{A similar objective can be defined as $S$ over edges $\E$. We define it over nodes as nodes often contain richer features than edges and are more flexible. One advantage of this choice will be made clear in Section \ref{subsec:evaluation}} is given by
\begin{equation} \label{eq:relaxed_obj}
    S^* = \argmax_{S \subseteq \V, |S| \leq \gamma|\V|} \sum_{i \in S}\score(f (\cdot), \gG, i)
\end{equation}


Existing methods often boil down to Equation \ref{eq:relaxed_obj} with different scoring functions (\score), and finding a proper \score is non-trivial. One example of \score is to evaluate each node $i$ directly as $\score(f (\cdot), \gG, i) = [f(\{i\})]_{c^*}$. However, this choice misses interactions between nodes and corresponds to a trivial case in GNNs where no message-passing is performed for $\{ i \}$. Another possibility is to use \eval as \score, e.g., $\score(f (\cdot), \gG, i) = \left[ f(\gG) \right]_{c^*} - \left[ f(\gG \backslash \{ i\}) \right]_{c^*}$. However, this again fails to capture interactions between nodes; for example, two nodes $i$ and $j$ may be both important but also complimentary, so their contribution to $\gG$ can only be observed when they are missing simultaneously. 

\subsection{Scoring functions from cooperative games} \label{subsec:motivation}
Given the challenges for defining a proper \score, solutions to cooperative games, like the Shapley value, have been proposed with $f (\cdot)$ as the characteristic function, i.e. $\score(f (\cdot), \gG, i) = \phi_i(|\gG|, f(\cdot))$ \cite{subgraphx, graphsvx}. However, existing works only use the non-structure-aware Shapley value. In contrast, values defined on games $(N, v, \gG)$ with communication structures $\gG$ are naturally structure-aware but were never considered GNN explanation. Below we discuss the non-structure-aware limitation of the Shapley value in detail and motivating structure-aware values with practical examples in GNN explanation.

The Shapley value is defined on games $(N, v)$, which by definition takes no graph structures. It assumes flexible cooperation between players and uniform distribution of coalition importance that only depends on $|S|$ (see Equation \ref{eq:shapley}). Even if a $\gG$ is given and the game is defined as $(N, v, \gG)$, the Shapley value will overlook $\gG$ when aggregating $m(i, S)$. In contrast, structure-aware values on $(N, v, \gG)$ can be interpreted as a weighted aggregation of coalitions with more reasonable weights. Although different solutions $\phi(N, v, \gG)$ have their nuances in weight adjustments \cite{hamiache_value, hn_value, myerson, d_myerson}, they share two key properties: \textbf{(1)} the weight is zero if $i$ and $S$ are disconnected because they are interpreted as players without communication channels \cite{myerson}, and \textbf{(2)} the weight is impacted by the nature of connections between $i$ and $S$ because it is easier for better-connected nodes to communicate. 

\noindent\textbf{A synthetic example.} We take the HN value (definition in Section \ref{subsec:hn_value}) as an example structure-aware value and compare it to the Shapley value in a simple graph in Figure \ref{figure:motivation_examples}\textbf{(a)}. To compute $\phi_1(N, v, \gG)$, both values aggregates $m(1, S)$ for $S \in \{\emptyset, \{2\}, \{ 3\}, \{2, 3\} \}$. The Shapley value first assigns a uniform weight $\frac{1}{3}$ to three different $|S|$, and then splits weights uniformly for the $|S| = 1$ case to be $\frac{1}{6}$. However, the HN value assigns weight zero for $S = \{3\}$ because 1 and 3 are disconnected in coalition $\{1, 3 \}$ and are assumed to be two independent graphs that shouldn't interact (property \textbf{(1)}). Their interaction is rather captured in the $S = \{2, 3\}$ case, when 1 and 3 are connected by the bridging node 2, and this case is also downweighted from $\frac{1}{3}$ to $\frac{1}{4}$, as 3 is relatively far from 1 (property \textbf{(2)}).

\noindent\textbf{A practical example.} The good properties of structure-aware values can help explain graph tasks. The example in Figure \ref{figure:motivation_examples}\textbf{(b)} is from \sst{} (dataset description in Section \ref{subsec:dataset}), where the graph for sentiment classification is constructed from the sentence \textit{``is still quite good-natured and not a bad way to spend an hour''} with edges generated by the Biaffine parser \cite{parser}. Assuming a model can correctly classify it as positive. Intuitively, ``good'' and ``not a bad'' are central to the human explanation. To compute the Shapley value of the word ``good'', the coalition ``not good'' will diminish the positive importance of ``good'', despite the two words lacking any direct connection. A structure-aware value can instead eliminate the \{``not'',  ``good''\} coalition, and only consider interactions between ``not'' and ``good'' (in fact, ``not'' and any other word) when the bridging ``bad'' appears, hence better binding ``not'' with ``bad'' and improving the salience of ``good''. In Section \ref{subsec:evaluation}, we revisit this example to observe impacts of structure-awareness empirically.

\section{GStarX: \texorpdfstring{\underline{G}raph \underline{St}ructure-\underline{a}wa\underline{r}e e\underline{X}planation}{Lg}}
\label{sec:methodology}
We propose \method{}, which uses a structure-aware HN-value-based \score to explain GNNs. We first state the definition of the HN value in cooperative game theory (\ref{subsec:hn_value}), and then connect it to the GNN message passing (\ref{subsec:gnn_and_hn}), and finally give the \method{} algorithm for GNN explanation (\ref{subsec:hn_explain}).
 
\subsection{The HN value} \label{subsec:hn_value}
Let $(N, v, \gG)$ be a game with a communication structure $\gG$ and $S \subseteq N$ be a coalition. Let $\bar S = \cup_{i\in S}\{\mathcal{N}(i) \}\cup S $ to be the union of $S$ and its neighbors in $\gG$. Let $S / \gG$ be the partition of $S$ containing connected components in $\gG$, i.e., $S / \gG = \{ \{ i \vert i = j \text{ or } i \text{ and } j \text{ are connected in \textit{S} by } \E \text{ of } \gG \} \vert j \in S\}$. Let $\gG[S]$ be the induced subgraph of $S$ in $\gG$. For example, in Figure \ref{figure:motivation_examples}\textbf{(b)}, when $S=$\{``is'', ``an'', ``hour''\}, $\bar S$ will be \{``is'', ``good'', ``an'', ``hour'', ``spend''\}, $S/ \gG$ will be \{\{``is''\}, \{``an'', ``hour''\}\}, and $\gG[S]$ will be the subgraph with a two-node component \ovalbox{an}-\ovalbox{hour} and a single node component \ovalbox{is}.

\begin{definition}[\textbf{Surplus}] \label{def:surplus}
The surplus $p(j, S)$ generated by a coalition $S$ cooperating with its neighbor $j$ is defined as
\begin{equation}
p(j, S) = v(S \cup \{ j \}) - v(S) - v(\{j \})
\end{equation}
\end{definition}
Intuitively, $p(j, S)$ is generated because $S$ is actively cooperating. Thus, when evaluating a fair payoff to $S$, a portion of $p(j, S)$ should be added to its own payoff $v(S)$. This idea leads to the next definition of associated games regarding the original games, where surplus allocation is performed.

\begin{definition}[\textbf{HN Associated Game}] \label{def:associated_game}
Given $0 \leq \tau \leq 1$ representing the portion of surplus that will be allocated to a coalition $S$ for its cooperation with other players. The HN associated game $(N, v_{\tau}^{*}, \gG)$ of $(N, v, \gG)$ is defined as 
\begin{numcases}{v_{\tau}^{*}(S) =}
    v(S) + \tau \sum_{j \in \bar S \backslash S}p(j, S)
         & \text{if } $|S/\gG| = 1$  \label{eq:associated_connected} \\ 
        \sum_{T \in S/\gG}v_{\tau}^{*}(T) 
        & otherwise \label{eq:associated_disconnected}
\end{numcases}
\end{definition}
The HN value is a solution on $(N, v, \gG)$. It is computed by iteratively constructing a series of HN associated games until it converges to a \textit{limit game} $(N, \tilde v, \gG)$. In other words, we first construct $v^*_{\tau}$ from $v$ by surplus allocation. Then we construct $v^{**}_{\tau}$ from $v^*_{\tau}$ by allocating the surplus generated from the $v^*_{\tau}$ and so on. The convergence of the limit game is guaranteed and the result $\tilde v$ is independent of $\tau$ under mild conditions as shown in \cite{hn_value}. The HN value of each player is uniquely determined by applying $\tilde v$ to that player, i.e. $\phi_i(N, v, \gG) = \tilde v (\{ i\})$. We state the formal definitions of the limit game and the uniqueness theorem of the HN value in Appendix \ref{app:limit_game}.

\subsection{Connecting GNNs and the HN surplus allocation through the message passing lens} \label{subsec:gnn_and_hn}
Both the GNN message passing (MP) and the associated game surplus allocation (SA) are iterative aggregation algorithms, with considerable alignment. In fact, SA on each singular node set $S = \{ i\}$ is exactly MP: Equation \ref{eq:associated_connected} becomes an instantiation of Equation \ref{eq:message_passing} with $\aggr(a, \vb) = a + \tau \sum_j \vb_j$ on a scalar node value $a$ and a neighbor set $\vb$. These algorithms differ in that SA applies more broadly to $|S|{\geq}1$ cases; it treats $S$ as a supernode when nodes in $S$ form a 
connected component in $\gG$, and handles disconnected $S$ component-wise via Equation \ref{eq:associated_disconnected}.

We illustrate SA using a real chemical graph example. The molecule shown in Figure \ref{figure:motivation_examples}\textbf{(c)} is taken from \mutag{} (dataset description in Section \ref{subsec:dataset}). It is known to be classified as \textit{mutagenic} because of the -NO2 group (nodes 1, 2, and 3) \cite{mutag}. When we compute $v^*_\tau (\{ 1\})$, the surplus $p(2, \{1\})$, $p(3, \{1\})$, and $p(4, \{1\})$ are allocated to node 1 (like messages passed to a central node in GNN). Then surplus are aggregated together with $v(\{1\})$ following Equation \ref{eq:associated_connected} to form $v_{\tau}^{*}(\{1\})$. 

For graphs, the SA approach has two advantages over the uniform aggregation approach used in the Shapley value: {\bf (1)} The aggregated payoff in each $v^*_{\tau}$ is structure-aware, like representations learned by GNNs \cite{gnn_structure_count}, and {\bf (2)} the iterative computation preserves locality, which is preserved by GNNs \cite{graph_locality}. In other words, these two properties mean close neighbors heavily influence each other due to cooperation in many iterations, while far away nodes less influence each other due to little cooperation. In the \mutag{} example, since the local -NO2 generates a high payoff for the mutagenicity classification, locally allocating the payoff helps us better understand the importance of the nitrogen atom and the oxygen atoms. Whereas aggregating over many unnecessary coalitions with far-away carbon atoms can obscure the true contribution of -NO2. We will revisit this example in Section \ref{subsec:evaluation}.

\subsection{The GStarX algorithm} \label{subsec:hn_explain}
We now state our algorithm for explaining GNNs with \method. Notice that \method{} scores nodes in a graph but not each dimension of node features. Feature dimension importance explanation is an orthogonal perspective that can be added on top of \method{}. We leave this extension as a future work. \method{} 
formulates the GNN explanation problem as a feature importance scoring problem, where nodes are scored to find the optimal node-induced subgraph as we introduced in Section \ref{subsec:feature_importance}. It essentially implements and solves the objective in Equation \ref{eq:relaxed_obj}, where an HN-value-based \score is used. To use such \score, we need to define the players and the characteristic function of the game, and then apply the formula in Equation \ref{eq:associated_connected} and \ref{eq:associated_disconnected}. Suppose the inputs are a graph $\gG$ with nodes $\V = \{u_1, \dots, u_n \}$ and label $y \in \{1, \dots, C\}$, a GNN $f(\cdot)$ outputs a probability vector $f(\gG)\in [0,1]^C$, and the predicted class $c^*=\argmax_{c} [f(\gG)]_{c}$. Let $\V$ be players, and let the normalized probability of the predicted class be the characteristic function $v$: 
\begin{equation} \label{eq:feature_score}
    v(S) = \left[ f(\gG[S]) \right]_{c^*} - f^0_{c^*} \quad \forall S \subseteq \V
\end{equation}
Here the normalization term $f^0_{c^*} = \mathbb{E}\left[ \left[ f(G) \right]_{c^*} \right]$ is the expectation over a random variable $G$ representing a general graph. In practice, we approximate it using the empirical expectation over all $\gG$ in the dataset. \score will be the HN value of the game, i.e., $\score(f (\cdot), \gG, i) = \phi_i(\V, v, \gG) = \tilde{v}(\{i\})$.


\begin{figure}[t] 
\begin{minipage}{0.58\textwidth}
\begin{algorithm}[H]
  \caption{\method: Graph Structure-Aware Explanation}
  \label{alg:main}
\begin{algorithmic}
  \STATE {\bfseries Input:} Graph $\gG$ with nodes $\V = \{u_1, \dots, u_n \}$, trained GNN $f(\cdot)$, empirical expectation $f^0$, hyperparameter $\tau$, max sample size $m$, number of samples $J$, sparsity $\gamma$.
  \STATE Get the predicted class $c^*=\argmax_{c}[f(\gG)]_{c}$ 
  \STATE Define characteristic function $v(S) = \left[ f(g_S) \right]_{c^*} - f^0_{c^*}$
\IF{$n \leq m$}
    \STATE $\phi = \computehn(\gG, \V, v(\cdot), \tau)$
\ELSE
    \STATE $\phi = \computehnmc(\gG, \V, v(\cdot), \tau, m, J) $
\ENDIF
\STATE Sort $\phi$ in descending order with indices $\{ \pi_1, \dots, \pi_n \}$
\STATE $k = \floor{\gamma |\V|}$
\STATE \textbf{Return:} $S^* = \{ u_{\pi_1}, \dots, u_{\pi_k} \}$
\end{algorithmic}
\end{algorithm}
\end{minipage}
\hfill
\begin{minipage}{0.4\textwidth}
\begin{algorithm}[H]
  \caption{The \computehn Function}
  \label{alg:computehn}
\begin{algorithmic}
  \STATE {\bfseries Input:} Graph instance $\gG$ with nodes $\V = \{u_1, \dots, u_n \}$, characteristic function $v$, hyperparameter $\tau$.
  \FOR{$S$ {in} $2^N$}
    \STATE Compute payoff $v(S)$ \COMMENT{Eq.(\ref{eq:feature_score})}
  \ENDFOR
  \STATE Construct matrix $\mH_{\{\tau, n, \gG\}}$ \COMMENT{Eq.(\ref{eq:matrix_form})}
  \REPEAT 
  \STATE $\mH = \mH  \mH$
  \UNTIL{$\mH$ converges} 
  \STATE Get the limit game $\tilde v = \mH v$ \COMMENT{Eq.(\ref{eq:limit_game})}
  \STATE Assign the first $n$ entries of $\tilde v$ to $\phi$ 
  \STATE \textbf{Return:} $\phi$
\end{algorithmic}
\end{algorithm}
\end{minipage}
\vskip -0.12in
\end{figure}

Given \score, we solve the objective by first computing the scores $\phi \in \R^n$ then selecting the top $\floor{\gamma |\V|}$ scores greedily as in Algorithm \ref{alg:main}. Practically, like other game-theoretic methods, the exact computation of the HN value is infeasible when the number of players $n$ is large. We thus do an exact computation for small graphs (the if-branch) and Monte-Carlo sampling for large graphs (the else-branch). The \computehn function is shown in Algorithm \ref{alg:computehn}, where the $\mH$ stands for a matrix form of the associated game defined in Definition \ref{def:associated_game}.(See Appendix \ref{app:limit_game} and \ref{app:hn_compute} for details of the matrix form and algorithms for \computehnmc). Also, even though the algorithm is stated for graph classification, \method{} works for node classification as well. This can be easily seen since GNNs classify nodes $u_i$ by processing an ego-graph centered at $u_i$, so the task can be converted to graph classification with the label of $u_i$ used as the label of the ego-graph. We focus on graph classification in the main text for simpler illustration and discuss more about node classification in Appendix \ref{app:node_classification}.

\section{Experiments} \label{sec:experiment}

\subsection{Experiment settings} \label{subsec:dataset}
\noindent\textbf{Datasets.} We conduct experiments on datasets from different domains including synthetic graphs, chemical graphs, and text graphs. A brief description of the datasets is shown below with more detailed statistics in Appendix \ref{app:dataset}

\begin{compactitem}[\textbullet]
\item \textbf{Chemical graph property prediction.} 
\mutag{} \cite{mutag}, \bace{} and \bbbp{} \cite{bbbp} contain chemical molecule graphs for graph classification, with atoms as nodes, bonds as edges, and chemical properties as graph labels.

\item \textbf{Text graph sentiment classification.} 
\sst{} and \twitter{} \cite{taxonomy} contain graphs constructed from text. Nodes are words with pre-trained BERT embeddings as features. Edges are generated by the Biaffine parser \cite{parser}. Graphs are labeled as positive or negative sentiment.

\item \textbf{Synthetic graph motif detection.}
\bamotif{} \cite{pgexplainer} contains graphs with a Barabasi-Albert (BA) base graph of size 20 and a 5-node motif in each graph. Node features are 10-dimensional all-one vectors. The motif can be either a house-like structure or a cycle. Graphs are labelled in two classes based on which motif they contain.
\end{compactitem}

\noindent\textbf{GNNs and explanation baselines.} We evaluate \method{} by explaining GCNs \cite{gcn} on all datasets in our major experiment in Section \ref{subsec:evaluation}. In the ablation study in Section \ref{subsec:analysis}, we further evaluate on GIN \cite{gin} and GAT \cite{gat} on certain datasets following \cite{subgraphx}. All models are trained to convergence with hyperparameters and performance shown in Appendix \ref{app:model}. We compare with 5 strong baselines representing the SOTA methods for GNN explanation: GNNExplainer \cite{gnnexplainer}, PGExplainer \cite{pgexplainer}, SubgraphX \cite{subgraphx}, GraphSVX \cite{graphsvx}, and OrphicX \cite{orphicx}. In particular, SubgraphX and GraphSVX use Shapley-value-based scoring functions.

\noindent\textbf{Evaluation metrics.} Evaluating explanations is non-trivial due to the lack of ground truth. We follow \cite{subgraphx, taxonomy} to employ \fidelity, Inverse Fidelity (\invfidelity), and \sparsity as our evaluation metrics. \fidelity and \invfidelity measure whether the prediction is faithfully important to the model prediction by removing the selected nodes or only keeping the selected nodes respectively. \sparsity promotes fair comparison by controlling explanations to have similar sizes, since including more nodes generally improves \fidelity and \invfidelity, and explanations with different sizes are not directly comparable. Ideal explanations should have high \fidelity, low \invfidelity, and high \sparsity, indicating relevance and conciseness. Equations \ref{eq:fidelity}-\ref{eq:sparsity} show their formulas.

\begin{equation}
    \fidelity(\gG, g) = \left[ f(\gG) \right]_{c^*} - \left[ f(\gG \backslash g) \right]_{c^*} \label{eq:fidelity}
\end{equation}
\begin{equation}
    \invfidelity(\gG, g) = \left[ f(\gG) \right]_{c^*} - \left[ f(g) \right]_{c^*} \label{eq:inv_fidelity} 
\end{equation}
\begin{equation}
     \sparsity(\gG, g) = 1 - {|g|}/{|\gG|} \label{eq:sparsity}
\end{equation}




\fidelity and \invfidelity are complementary and are both important for a good explanation $g$. \fidelity justifies the necessity for $g$ to be included to predict correctly. \invfidelity justifies the sufficiency of a standalone $g$ to predict correctly. As they are analogous to precision and recall, we draw an analogy to the F1 score to propose a single-scalar-metric ``harmonic fidelity'' (\harfidelity), where we normalize them by \sparsity and take their harmonic mean; see Appendix \ref{app:metric} for the formula.


\noindent\textbf{Hyperparameters.} 
\method{} includes three hyperparameters: $\tau$ for the allocated surplus in the associated game,  $m$ as the maximum graph size to perform exact HN value calculation, and $J$ as the number of samples for the MC approximation. In our experiments, we choose $\tau = 0.01$ since we need $\tau < \frac{2}{n}$ for convergence (Appendix \ref{app:limit_game}) and all graphs in the datasets above have less than 200 nodes. For $m$ and $J$, bigger values should be better for the MC approximation, and we found $m=10$ and $J=n$ work well empirically.

\begin{figure*}[t]
\begin{center}
\includegraphics[width=\textwidth]{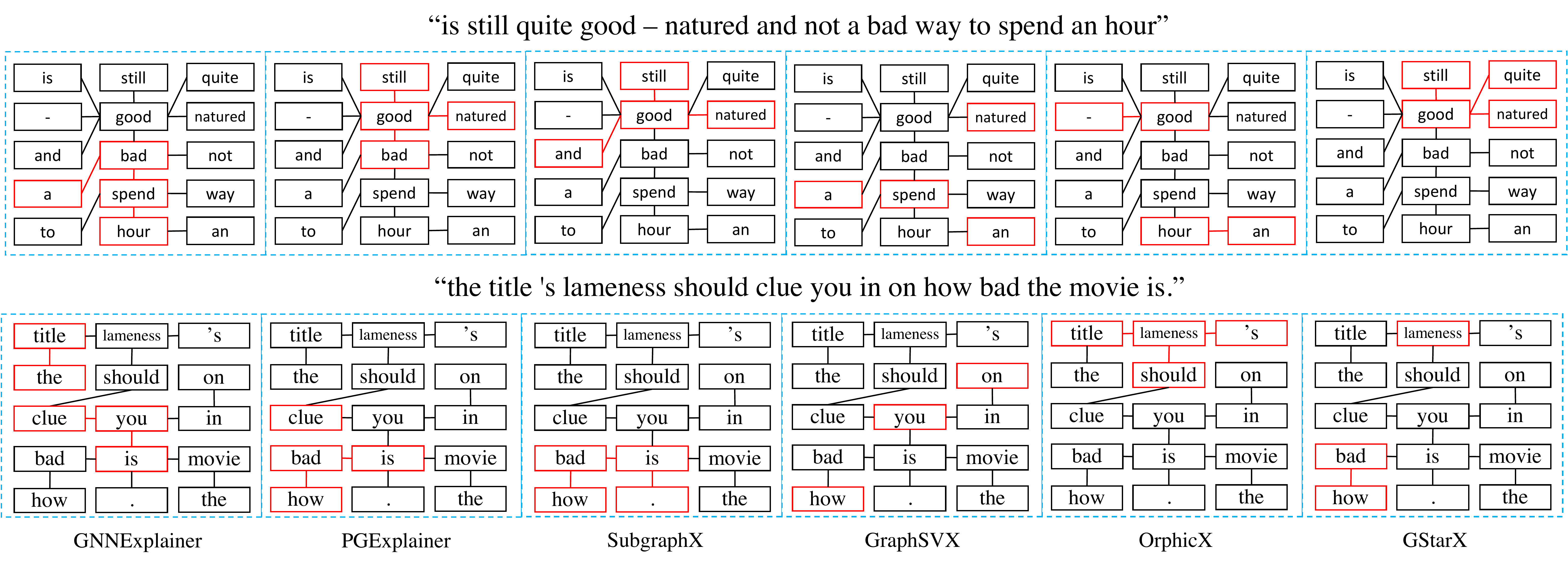}
\end{center}
\caption{Explanations on sentences from \sst{}. We show the explanation of one positive sentence (upper) and one negative sentence (lower). Red outlines indicate the selected nodes/edges as the explanation. \method{} identifies the sentiment words more accurately compared to baselines.}
\label{figure:sst_examples}
\vskip -0.05in
\end{figure*}

\begin{figure*}[t]
\begin{center}
\includegraphics[width=0.8\textwidth]{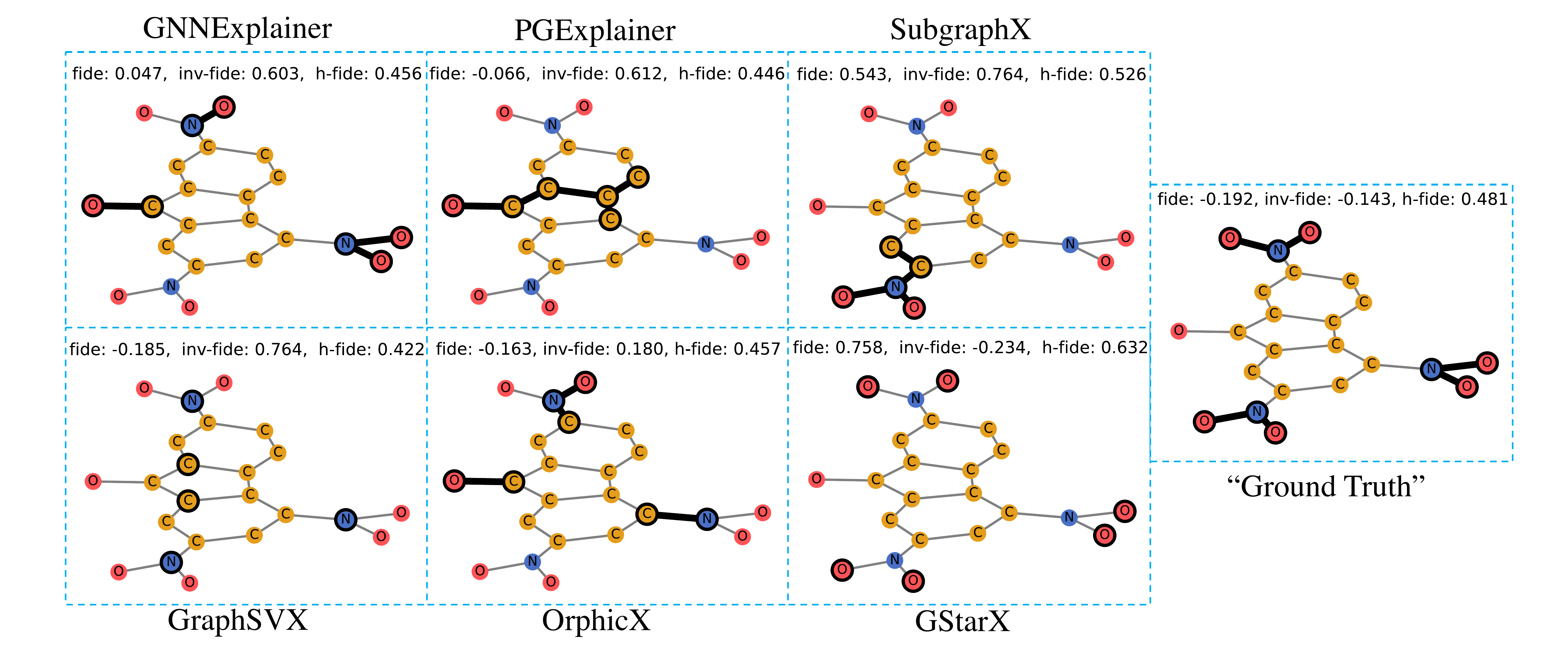}
\end{center}
\caption{Explanations on a mutagenic molecule in \mutag{}. Carbon atoms (C) are in yellow, nitrogen atoms (N) are in blue, and oxygen atoms are in red (O). Dark outlines indicate the selected nodes/edges as the explanation. We report the explanation \fidelity (fide), \invfidelity (inv-fide), and \harfidelity (h-fide). \method{} gives a significantly better explanation than other methods in terms of these metrics.}
\label{figure:mutag_examples}
\vskip -0.2in
\end{figure*}

\subsection{Evaluation results} \label{subsec:evaluation}
\noindent\textbf{Quantitative studies.} We report averaged test set \harfidelity in Table \ref{tab:harmonic_fidelity}. We conduct 8 different runs to get results with \sparsity ranging from 0.5-0.85 in 0.05 increments (\sparsity cannot be precisely guaranteed, hence it has minor variations across methods) and report the best \harfidelity for each method. \method{} outperforms others on 4/6 datasets and has the highest average. We also follow \cite{subgraphx} to show the \fidelity vs. \sparsity plots for all 8 sparsity in Appendix \ref{app:f_vs_sp}.

\begin{table}[t]
\center
\small
\caption{The best \harfidelity (higher is better) of 8 different \sparsity for each dataset. \method{} shows higher \harfidelity on average and on 4/6 datasets.}
\begin{tabular}{@{}lllllll@{}}
\toprule
\multicolumn{1}{l}{Dataset} & \multicolumn{1}{l}{GNNExplainer} & PGExplainer & SubgraphX & GraphSVX & OrphicX & \method{}  \\ \midrule
\bamotif & 0.4841 & 0.4879 & \textbf{0.6050} & 0.5017 & 0.5087 & 0.5824  \\
\bace & 0.5016 & 0.5127 & 0.5519 & 0.5067 & 0.4960 & \textbf{0.5934}   \\ 
\bbbp & 0.4735 & 0.4750 & \textbf{0.5610} & 0.5345 & 0.4893 & 0.5227   \\ 
\sst & 0.4845 & 0.5196 & 0.5487 & 0.5053 & 0.4924 & \textbf{0.5519}    \\ 
\mutag & 0.4745 & 0.4714 & 0.5253 & 0.5211 & 0.4925 & \textbf{0.6171}  \\
\twitter & 0.4838 & 0.4938 & 0.5494 & 0.4989 & 0.4944 & \textbf{0.5716} \\ \midrule
Average & 0.4837 & 0.4934 & 0.5569 & 0.5114 & 0.4952 & \textbf{0.5732} \\ 
\bottomrule
\end{tabular}
\label{tab:harmonic_fidelity}
\vskip -0.15in
\end{table}

\noindent\textbf{Qualitative studies.} We visualize the explanations of graphs in \sst{} in Figure \ref{figure:sst_examples} and compare them qualitatively. We show explanations selected with high and comparable \sparsity on a positive (upper) graph and a negative (lower) graph. \method{} concisely captures the important words for sentiment classification without including extraneous ones for both sentences. Baseline methods generally select some-but-not-all important sentiment words, with extra neutral words as well. Among baselines, SubgraphX gives more reasonable results. However, it cannot cover two groups of important nodes with a limited budget because it can only select a connected subgraph as the explanation; e.g. to cover the negative word ``lameness'' in the lower sentence, SubgraphX needs at least three more nodes along the way, which will significantly decrease \sparsity while including undesirable, neutral words. Moreover, we discussed in Section \ref{subsec:motivation} that the Shapley value will downgrade the positive importance of the word “good” for the upper sentence. Comparing the normalized contribution scores of our HN-value-based method \method{} and the Shapley-based method GraphSVX, contribution of ``good'' is higher in ours: 0.1152 vs. 0.0371.

We visualize explanations selected with high and comparable \sparsity of a mutagenic molecule from \mutag{} in Figure \ref{figure:mutag_examples}. Explanations on chemical graphs are harder to evaluate than text graphs as they require domain knowledge. \mutag{} has been widely used as a benchmark for evaluating GNN explanations because human experts recognize -NO2 as mutagenic \cite{mutag}, which makes \mutag{} a dataset with ``ground truth''\footnote{Carbon rings were also claimed as mutagenic by human experts, but we found it is not discriminative as they exist in both mutagenic and non-mutagenic molecules in \mutag{}.}. Surprisingly, we found that \method{} generates much better \harfidelity/\fidelity/\invfidelity than other methods and even the ``ground truth'' by only selecting the -O in -NO2 as explanations. In particular, the -0.234 \invfidelity of \method{} means the selected subgraph has an even better prediction result than the original whole graph (0 \invfidelity) and the ground truth (-0.143 \invfidelity) because nodes not significant to the GNN prediction are removed. Fidelity metrics of baselines are inferior to \method{} because they include other non-discriminative carbon atoms despite they capture -NO2 to some extent. This suggests that even though \textbf{human experts identify -NO2} as the ``ground truth'' of mutagenicity, the \textbf{GNN only needs -O} to classify mutagenic molecules. With the goal being understand model behavior, \method{} explanation is better. Moreover, SubgraphX is the only baseline that has better \harfidelity than the ``ground truth'', but it can only capture one -NO2 because its search algorithm requires the explanation to be connected, so its \invfidelity is not optimal. In fact, GNNExplainer, PGExplainer, and SubgraphX can never generate explanations including only disconnected -O without -N like \method, because the former two solve the explanation problem by optimizing edges (as opposed to Equation \ref{eq:relaxed_obj}), and the latter requires connectedness. More \mutag{} explanation visualizations are in Appendix \ref{app:visualization}.

\subsection{Ablation study and analysis} \label{subsec:analysis}
\noindent\textbf{Model-agnostic explanation.} \method{} makes no assumptions about the model architecture and can be applied to explain various GNN backbones. We use GCN for all datasets in the major experiment above for consistency, and we now further investigate performance on two more popular GNNs: GIN and GAT. We follow \cite{subgraphx} to train GIN on \mutag{} and GAT on \sst \footnote{As some baselines take over $24$ hours on full \sst{}, we randomly select 30 graphs for this analysis.}, and show results in Table \ref{tab:model_agnostic}. For both settings, \method{} outperforms the baselines, which is consistent with results on GCN.

\noindent\textbf{Efficiency study.} The \method{} algorithm scales in $O(J)$ with practical $J \propto |\V|$. Following \cite{subgraphx}, we study the empirical efficiency of \method{} by explaining 50 randomly selected graphs from \bbbp{}. We report the average run time in Table \ref{tab:efficiency}. Our results for the baselines are similar to \cite{subgraphx}. \method{} is not the fastest method, but it is more than two times faster than SubgraphX. Since explanation usually doesn't have strict efficiency requirements in real applications, considering \method{} generates higher-quality explanations than the baselines, we believe the time complexity of \method{} is acceptable.

\begin{table}[t]
\center
\small
\caption{\method{} shows higher \harfidelity for both GAT on \sst{} and GIN on \mutag{}.}
\begin{tabular}{@{}lllllll@{}}
\toprule
\multicolumn{1}{l}{Dataset} & \multicolumn{1}{l}{GNNExplainer} & PGExplainer & SubgraphX & GraphSVX & OrphicX & \method{} \\ \midrule
\sst\xspace & 0.4951 & 0.4918 & 0.5484 & 0.5132 & 0.4997 & \textbf{0.5542}    \\ 
\mutag\xspace & 0.5042 & 0.4993 & 0.5264 & 0.5592 & 0.5152 & \textbf{0.6064}  \\
\bottomrule
\end{tabular}
\label{tab:model_agnostic}
\vskip -0.2in
\end{table}

\begin{table}[t]
\center
\small
\caption{Average running time on 50 graphs in \bbbp}
\begin{tabular}{@{}lllllll@{}}
\toprule
\multicolumn{1}{l}{Method} & \multicolumn{1}{l}{GNNExplainer} & PGExplainer & SubgraphX & GraphSVX & OrphicX & \method{}  \\ \midrule
Time(s) & 11.92 & 0.03 (train 720) & 75.96 & 3.06 & 0.15 (train 915) & 31.24    \\ 
\bottomrule
\end{tabular}
\label{tab:efficiency}
\end{table}

\noindent\textbf{Explanation sparsity study.} To further study whether the obtained scores by \method{} are sparse, we follow \cite{zorro} to evaluate an entropy-based sparsity measure on model output scores. We show the average \method{} entropy-based sparsity on all datasets, and compare them with three reference score distributions on all $n$ nodes in a graph. 1) An upper bound: Uniform(n), which represents the least sparse output. 2) A practical lower bound: Uniform(0.25*n) which represents very sparse outputs with only top 25\% of nodes. 3) Poisson(0.25*n), which is a more realistic version of case 2). Results in Table \ref{tab:entropy_sparsity} show the average entropy-based sparsity of \method{} is much lower than Uniform(n) and close to Poisson(0.25*n), which justifies the \method{} outputs are indeed sparse. A more detailed discussion of this metric and these three reference distributions is in Appendix \ref{app:entropy_sparsity}.



\begin{table}[t]
\center
\small
\caption{The entropy-based sparsity scores of \method{} vs. three reference distributions, which shows \method{} outputs are indeed sparse.}
\begin{tabular}{@{}lllllll@{}}
\toprule
\multicolumn{1}{l}{Dataset} & \multicolumn{1}{l}{\bamotif} & \bace & \bbbp & \sst & \mutag & \twitter \\ \midrule
\method{} & 2.1352 & 2.4481 & 2.3290 & 2.3282 & 2.2434 & 2.2114   \\
Uniform(n) & 3.2189 & 3.5080 & 3.0728 & 2.8698 & 2.8612 & 2.9833   \\
Uniform(0.25*n) & 1.8326 & 2.1217 & 1.6893 & 1.4855 & 1.4749 & 1.5970 \\
Poisson(0.25*n) & 2.3204 & 2.4686 & 2.2416 & 2.1336 & 2.1323 & 2.1945 \\
\bottomrule
\end{tabular}
\label{tab:entropy_sparsity}
\vskip -0.2in
\end{table}


\section{Related work}\label{sec:related}
\noindent\textbf{GNN explanation} 
aims to produce an explanation for a GNN prediction on a given graph, usually as a subgraph induced by important nodes or edges. Many existing methods work by scoring nodes or edges and are thus similar to this work. For example, the scoring function of GNNExplainer \cite{gnnexplainer} is the mutual information between a masked graph and the prediction on the original graph, where soft masks on edges and node features are generated by direct parameter learning. PGExplainer \cite{pgexplainer} uses the same scoring function as \cite{gnnexplainer} but generates a discrete mask on edges by training an edge mask predictor. SubgraphX \cite{subgraphx} uses the Shapley value as its scoring function on subgraphs selected by Monte Carlo Tree Search (MCTS), and GraphSVX \cite{graphsvx} uses a least-square approximation to the Shapley value to score nodes and their features. While SubgraphX and GraphSVX were shown to perform better than prior alternatives, as we show in Section \ref{sec:motivation}, the Shapley value they try to approximate is non-ideal as it is non-structure-aware. Although SubgraphX and GraphSVX use $L$-hop subgraphs and thus technically they use the graph structure, such structure usage are very limited in achieving structure-awareness as we show in Appendix \ref{app:cutoff}. While there are many other GNN explanation methods from very different perspectives, i.e. gradient analysis \cite{cam}, model decomposition \cite{lrp}, surrogate models \cite{pgm_explainer}, and causality \cite{gem, orphicx}, we defer their details to Appendix \ref{app:more_related} given their lesser relevance. 

\noindent\textbf{Cooperative game theory} originally studies how to allocate payoffs among a set of players in a cooperative game. Recently, certain ideas from this domain have been successfully used in feature importance scoring for ML model explanation \cite{shapley_regression, shapley_sampling, SHAP}. When used for model explanation, data features becomes players in the game, e.g. pixels for images, and the value of the game gives feature importance scores. The vast majority of works in this line, like the ones cited above, deem the Shapley value \cite{shapley} to be the only choice. In fact, there are many other values with different properties and used in different situations in cooperative game theory. However, to the best of our knowledge, only \cite{l_and_c} mentions the Myerson value \cite{myerson} in the context of proposing a connected Shapley (C-Shapley) value for explaining sequence data, and it is not directly comparable to ours for graph data. A detailed discussion of the Myerson value and the C-Shapley value can be found in Appendix \ref{app:myerson_and_c_shapley}. Our work follows the cooperative game theory approach to explain models on graph data using the HN value \cite{hn_value}, which as we show is a better choice than the Shapley value given its structure-awareness.

\section{Conclusion and future work}\label{sec:conclusion}
In summary, we study GNN explanation on graphs via node importance scoring. We identify the non-structure-aware challenge of existing Shapley-value-based approaches and propose \method{} to assign importance scores to each node via a structure-aware HN value. We also build connections between the HN value surplus allocation and GNN message passing. GStarX demonstrates its superiority over strong baselines on chemical and text graph classifications. A limitation of \method{} is that the importance of different node feature dimensions is not explained. One future work is to add this extension, which could be done by scoring a subset of nodes together with a subset of features each time. Another future direction is to exploit the rich cooperative game theory literature. Beyond the Shapley value, more values are possible for explaining ML models. For graph data, edge-based values like \cite{position_value} can potentially be applied to an alternative edge-based objective like Equation \ref{eq:relaxed_obj}. Other values may be appropriate to more data types beyond graphs.

\section*{Acknowledgement}
This work was partially supported by NSF III-1705169, NSF 1937599, NSF 2119643, Okawa Foundation Grant, Amazon Research Awards, Cisco research grant USA000EP280889, Picsart Gifts, and Snapchat Gifts.

\bibliography{reference}
\bibliographystyle{plain}

\section*{Checklist}
\begin{enumerate}

\item For all authors...
\begin{enumerate}
  \item Do the main claims made in the abstract and introduction accurately reflect the paper's contributions and scope?
    \answerYes{}
  \item Did you describe the limitations of your work?
    \answerYes{}
  \item Did you discuss any potential negative societal impacts of your work?
    \answerNo{}
  \item Have you read the ethics review guidelines and ensured that your paper conforms to them?
    \answerYes{}
\end{enumerate}

\item If you are including theoretical results...
\begin{enumerate}
  \item Did you state the full set of assumptions of all theoretical results?
    \answerNA{}
        \item Did you include complete proofs of all theoretical results?
    \answerNA{}
\end{enumerate}

\item If you ran experiments...
\begin{enumerate}
  \item Did you include the code, data, and instructions needed to reproduce the main experimental results (either in the supplemental material or as a URL)?
    \answerYes{}
  \item Did you specify all the training details (e.g., data splits, hyperparameters, how they were chosen)?
    \answerYes{}
        \item Did you report error bars (e.g., with respect to the random seed after running experiments multiple times)?
    \answerYes{}
        \item Did you include the total amount of compute and the type of resources used (e.g., type of GPUs, internal cluster, or cloud provider)?
    \answerYes{}
\end{enumerate}

\item If you are using existing assets (e.g., code, data, models) or curating/releasing new assets...
\begin{enumerate}
  \item If your work uses existing assets, did you cite the creators?
    \answerYes{}
  \item Did you mention the license of the assets?
    \answerYes{}
  \item Did you include any new assets either in the supplemental material or as a URL?
    \answerNo{}
  \item Did you discuss whether and how consent was obtained from people whose data you're using/curating?
    \answerNA{}
  \item Did you discuss whether the data you are using/curating contains personally identifiable information or offensive content?
    \answerNA{}
\end{enumerate}

\item If you used crowdsourcing or conducted research with human subjects...
\begin{enumerate}
  \item Did you include the full text of instructions given to participants and screenshots, if applicable?
    \answerNA{}
  \item Did you describe any potential participant risks, with links to Institutional Review Board (IRB) approvals, if applicable?
    \answerNA{}
  \item Did you include the estimated hourly wage paid to participants and the total amount spent on participant compensation?
    \answerNA{}
\end{enumerate}

\end{enumerate}

\clearpage
\appendix
\title{Explaining Graph Neural Networks with Structure-Aware Cooperative Games: Appendix}

\section{Experiment details}
\subsection{Dataset statistics} \label{app:dataset}
In Table \ref{tab:dataset}, we provided the statistics of all datasets used in our experiments.

\begin{table*}[ht]
\center
\small
\caption{Dataset Statistics.}
\begin{tabular}{@{}lllllll@{}}
\toprule
\multicolumn{1}{l}{Dataset} & \multicolumn{1}{l}{\# Graphs} & {\# Test Graphs} & \# Nodes (avg) & \# Edges (avg) & \# Features & \# Classes \\ \midrule
\mutag & 188 & 20 & 17.93 & 19.79 & 7 & 2  \\
\bace & 1,513 & 152 & 34.01 & 73.72 & 9 & 2   \\ 
\bbbp & 2,039 & 200 & 24.06 & 25.95 & 9 & 2   \\ 
\sst & 70,042 & 1821 & 9.20 & 10.19 & 768 & 2   \\ 
\twitter & 6,940 & 692 & 21.10 & 40.20 & 768 & 3   \\ 
\bamotif & 1,000 & 100 & 25 &  25.48 & 10  & 2 \\ 
\bottomrule
\end{tabular}
\label{tab:dataset}
\end{table*}

\subsection{Model architectures and implementation} \label{app:model}
In Table \ref{tab:gcn}, we provided the hyperparameters and test accuracy for the GCN model used in our major experiments. In Table \ref{tab:model_agnostic}, we provided the hyperparameters and test accuracy for the GIN and GAT model used in our analysis experiment. Most parameters are following \cite{subgraphx}, with small changes to further boost the test accuracy.

We run all experiments on a machine with 80 Intel(R) Xeon(R) E5-2698 v4 @ 2.20GHz CPUs, and a single NVIDIA V100 GPU with 16GB RAM. Our implementations are based on Python 3.8.10,  PyTorch 1.10.0, PyTorch-Geometric 1.7.1 \cite{pyg}, and DIG \cite{dig}. We adapt the GNN implementation and most baseline explainer implementation from the DIG library, except for GraphSVX and OrphicX where we adapt the official implementation. For the baseline hyperparameters, we closely follow the setting in \cite{subgraphx} and \cite{graphsvx} for a fair comparison. Please refer to \cite{subgraphx} Section 4.1 and \cite{graphsvx} Appendix E for details.

\begin{table}[H]
\center
\caption{GCN architecture hyperparameters according to results in Table \ref{tab:gcn}}
\begin{tabular}{@{}llllll@{}}
\toprule
\multicolumn{1}{l}{Dataset} & \multicolumn{1}{l}{\#Layers} & \#Hidden & Pool & Test Acc  \\ \midrule
\bamotif & 3 & 20 & mean & 0.9800  \\ 
\bace & 3 & 128 & max & 0.8026 \\
\bbbp & 3 & 128 & max & 0.8634    \\ 
\mutag & 3 & 128 & mean& 0.8500   \\
\sst & 3 & 128 & max & 0.8808    \\ 
\twitter & 3 & 128 & max & 0.6908    \\ 
\bottomrule
\end{tabular}
\label{tab:gcn}
\end{table}

\begin{table}[H]
\center
\small
\caption{GIN and GAT architecture hyperparameters according to results in Table \ref{tab:model_agnostic}. For GAT, we use 10 attention heads with 10 dimension each, and thus 100 hidden dimensions.}
\begin{tabular}{@{}llllll@{}}
\toprule
\multicolumn{1}{l}{Dataset} & \multicolumn{1}{l}{\#Layers} & \#Hidden & Pool & Test Acc  \\ \midrule
\sst (GAT) & 3 & 10 \texttimes 10 & max & 0.8814 \\ 
\mutag (GIN) & 3 & 128 & max & 1.0   \\
\bottomrule
\end{tabular}
\label{tab:gat_gin}
\end{table}

\subsection{Exact formula for evaluation metrics} \label{app:metric}
Formulas for \fidelity, \invfidelity, and \sparsity are shown in Equation \ref{eq:fidelity}, \ref{eq:inv_fidelity}, and \ref{eq:sparsity}. In Equation \ref{eq:norm_fidelity}, \ref{eq:norm_inv_fidelity}, and \ref{eq:harmonic_fidelity}, we show formulas for normalized fidelity (\normfidelity), normalized inverse fidelity (\norminvfidelity), and harmonic fidelity (\harfidelity). Both the \normfidelity and \norminvfidelity are in $[-1, 1]$. The \harfidelity flips \norminvfidelity, rescales both values to be in $[0, 1]$, and takes their harmonic mean.

\begin{equation} \label{eq:norm_fidelity}
\normfidelity(\gG, g) = \fidelity(\gG, g) \cdot (1 - \frac{|g|}{|\gG|})
\end{equation}
\begin{equation} \label{eq:norm_inv_fidelity}
\norminvfidelity(\gG, g) = \invfidelity(\gG, g) \cdot (\frac{|g|}{|\gG|})
\end{equation}

Let $m1=\normfidelity(\gG, g)$, $m2=\norminvfidelity(\gG, g)$
\begin{align} \label{eq:harmonic_fidelity}
\harfidelity(\gG, g) 
&= \frac{2}{(\frac{1 + m1}{2})^{-1} + (\frac{1 - m2}{2})^{-1}} \nonumber \\
&= \frac{(1 + m1) \cdot (1 - m2)}{(2 + m1 - m2)}
\end{align}

\subsection{Fidelity vs. sparsity plots} \label{app:f_vs_sp}
\begin{figure*}[ht]
\begin{center} 
\includegraphics[width=\textwidth]{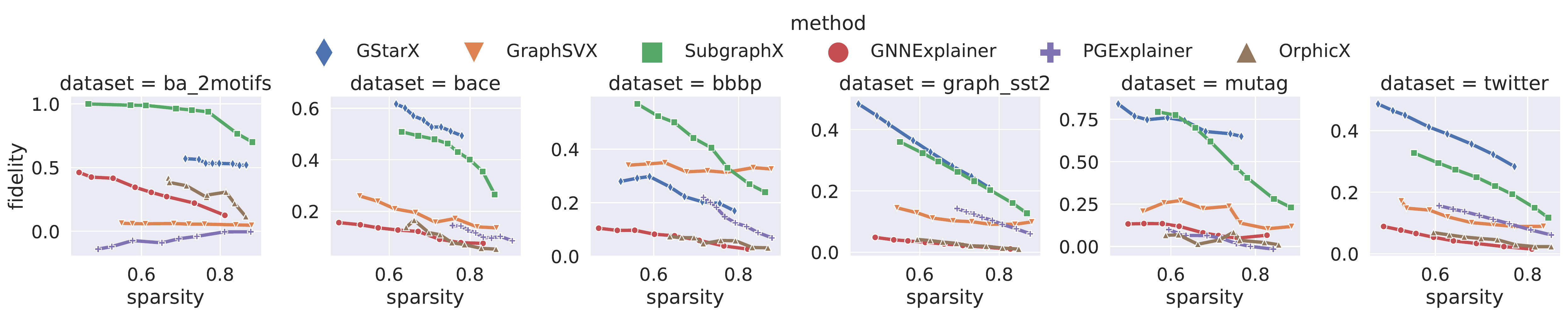}
\includegraphics[width=\textwidth]{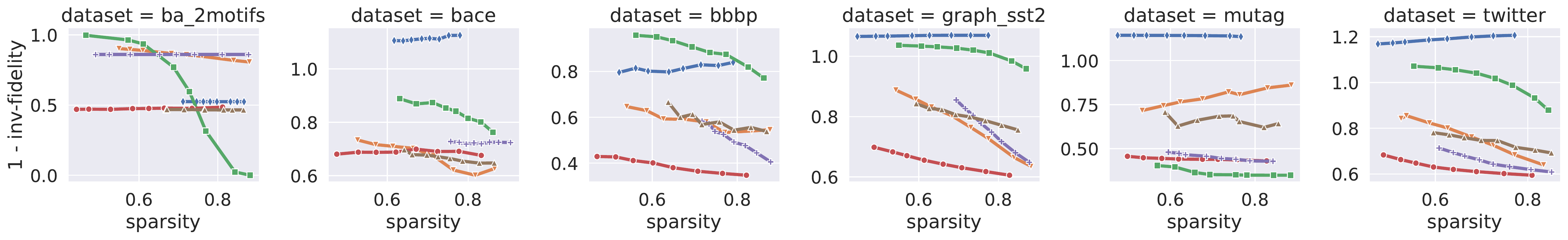}
\includegraphics[width=\textwidth]{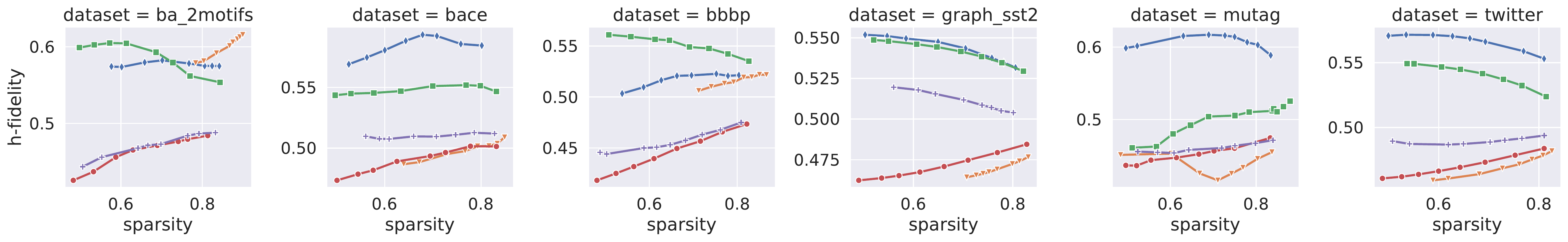}
\end{center}
\caption{\fidelity (row1), 1 - \invfidelity (row2), and \harfidelity (row3) vs. \sparsity on all datasets corresponding to the results shown in Table \ref{tab:harmonic_fidelity}. All three metrics are the higher the better. We see that \method{} outperforms the other methods}
\label{figure:f_vs_sp}
\end{figure*}

In Table \ref{tab:harmonic_fidelity}, we report the best \harfidelity among 8 different sparsities for each method on each dataset. We also follow \cite{subgraphx} to show the \fidelity vs. \sparsity plots in Figure \ref{figure:f_vs_sp} row1. Note that GraphSVX tends to give sparse explanations on some datasets, we still pick 8 different sparsities for it but mostly on the higher end. We also show the \textit{1 - \invfidelity} vs. sparsity plots and the \harfidelity vs. sparsity plots. Curves in all three plots are the higher the better.

\subsection{Detailed entropy-based sparsity evaluation} \label{app:entropy_sparsity}
In Section \ref{subsec:evaluation} we study whether the obtained scores by \method{} are sparse and follow \cite{zorro} to apply an entropy-based sparsity measure on scores. We now provide a more detailed discussion of this study.

The entropy-based sparsity, as defined in Definition 2 in \cite{zorro}, is shown in the Equation \ref{eq:entropy_sparsity} below. Here $\phi$ is the model output scores for a data instance, and $\tilde \phi_i = \frac{\phi_i}{\sum_{i} \phi_i}$ represent normalized scores.

\begin{equation} \label{eq:entropy_sparsity}
    H(\tilde \phi) = - \sum_{i \in n} \tilde \phi_i \log \tilde \phi_i
\end{equation}

The entropy-based sparsity helps us to understand how sparse an explanation is, before the scores are turned into hard explanation by thresholding or selecting top k. In Table \ref{tab:entropy_sparsity}, we show the average scores for \method{} on all datasets, and compare them with three reference cases. 1) The entropy of uniform distribution over all n nodes in a graph, i.e., Uniform(n), which represents the least sparse output and is an upper bound of entropy-based sparsity. 2) The entropy of uniform distribution over the top 25\% nodes in a graph, i.e., Uniform(0.25*n), where probabilities of the bottom 75\% nodes are set to zero. This case is very sparse since 75\% of nodes are deterministically excluded, which can be treated as a practical lower bound of entropy-based sparsity. 3) The entropy of Poisson distribution with mean 0.25*n, i.e. Poisson(0.25*n), which is a more realistic version of the sparse output in case 2). Instead of setting all 75\% of nodes to have probability zero, we assume the probabilities for tail nodes decrease exponentially as a Poisson distribution while the mean is kept the same as in case 2). Results in Table \ref{tab:entropy_sparsity} show that the average entropy-based sparsity of GStarX is between Uniform(0.25*n) and Uniform(n) and close to Poisson(0.25*n), which justifies the GStarX outputs are indeed sparse.

\section{GStarX for node classification}\label{app:node_classification}
Even though the \method{} algorithm is stated for graph classification, it works for node classification as well. This can be easily seen as the GNN node classification can be covert to classify an ego-graph. Given a graph $\gG$ with $\V = \{u_1, \dots, u_n \}$. Node classification on $u_i$ with an $L$-layer GNN can be converted to a graph classification. The target graph to classify will be the $L$-hop ego-graph centered at $u_i$, because this is the receptive field of the GNN for classifying $u_i$ and nodes further away won't influence the result. The label of the graph will be the label of $u_i$. In this case, the final readout layer of the GNN will be indexing $u_i$ instead of pooling. Given this kind of conversion, everything we showed in Section \ref{sec:methodology} follows.




\section{More related work}\label{app:more_related}
\paragraph{GNN explanation continued}
Besides the perturbation-based method we mentioned in Section \ref{sec:related}, there are several other types of approaches for GNN explanation. Gradient-based methods are widely used for explaining ML models on images and text. The key idea is to use the gradients as the approximations of input importance. Such methods as contrastive gradient-based (CG) saliency maps, Class Activation Mapping (CAM), and gradient-weighted CAM (Grad-CAM) have been generalized to graph data in \cite{cam}. Decomposition-based methods are a popular way to explain deep NNs for images. They measure the importance of input features by decomposing the model predictions and regard the decomposed terms as importance scores. Decomposition methods including Layer-wise Relevance Propagation (LRP) and Excitation Backpropagation (EB) have also been extended to graphs \cite{cam, lrp}. Surrogate-based methods work by approximating a complex model using an explainable model locally. Possible options to approximate GNNs include linear model as in GraphLIME \cite{graphlime}, additive feature attribution model with the Shapley value as in GraphSVX \cite{graphsvx}, and Bayesian networks as in \cite{pgm_explainer}. GNN explainability has also been studied from the causal perspective. In \cite{gem, orphicx}, generative models were constructed to learn causal factors, and explanations were produced by analyzing the cause-effect relationship in the causal graph.

\section{Properties of the Shapley value} \label{app:shapley}
The Shapley value was proposed as the unique solution of a game $(N, v)$ that satisfies three properties shown below, i.e. \textit{efficiency}, \textit{symmetry}, and \textit{additivity} \cite{shapley}. These three properties together are referred as an axiomatic characterization of the Shapley value. The \textit{associated consistency} properties introduced in Section \ref{subsec:hn_value} provides a different axiomatic characterization.

\property[\textbf{Efficiency}]{
\begin{equation*}
\sum_{i \in N} \phi_i(N, v)  = v(N) 
\end{equation*}
}

\property[\textbf{Symmetry}]{ If $v(S \cup \{i\}) = v(S \cup \{j\})$ for all $S \in N \backslash \{i, j\}$, then
\begin{equation*}
\phi_i(N, v) = \phi_j(N, v)
\end{equation*}
}

\property[\textbf{Additivity}]{ Given two games $(N, v)$ and $(N, w)$, 
\begin{equation*}
\phi(N, v + w) = \phi(N, v) + \phi(N, w)
\end{equation*}
}

The efficiency property states that the value should fully distribute the payoff of the game. The symmetry property states that if two players make equal contributions to all possible coalitions formed by other players (including the empty coalition), then they should have the same value. The additivity property states that the value of two independent games should be added player by player. It is the most useful for a system of independent games.

\section{Properties and calculation of the HN value}
\subsection{Consistency and associated games} \label{app:associated_game}
One reason for the Shapley value's popularity is its \textit{axiomatic characterization}, indicating that it is the unique solution that satisfies a set of desirable properties (see Appendix \ref{app:shapley}). Then \cite{associated_shapley} proposed a new axiomatic characterization of the Shapley value based on a different \textit{associated consistency} property. The \textit{consistency} property is a common analysis tool used in game theory \cite{potential, consistency1, consistency2, consistency3}.  The idea is to analyze a game $(N, v)$ by defining other reduced games $(S, v_S)$ for $S \subseteq N$, and a solution function $\phi$ is called \textit{consistent} when $\phi(N, v)$ yields the same payoff as $\phi(S, v_S)$ on each $S$. When $(S, v_S)$ is defined with desired properties, these good properties can be enforced for a solution by requiring consistency. The associated consistency in \cite{associated_shapley} is a special case of consistency between $(N, v)$ and only one other game $(N, v^*)$, which is called the \textit{associated game}. \cite{associated_shapley} shows that a carefully designed associated game uniquely characterizes the Shapley value. Associated consistency is also the key idea of the HN value.

\subsection{Limit game and the axiomatic characterization} \label{app:limit_game}
The HN value is established on a special associated game as we discussed in Section \ref{subsec:hn_value}. We can actually write this associated game in a more compact matrix form, where we slightly abuse notation and use $v$ and $v_{\tau}^{*}$ to represent vectors of payoffs for all $S \subseteq N$ under the original and associated game respectively. In other words, $v(S)$, which is used to represent evaluating the coalition $S$ using the characteristic function $v$, now can also be interpreted as indexing the vector $v$ with index $S$. 

\begin{lemma} \label{lemma:matrix_form}
A matrix form of the associated game $(N, v_{\tau}^{*}, \gG)$ is given by 
\begin{equation} \label{eq:matrix_form}
v_{\tau}^{*} = \mH_{\{\tau, n, \gG\}} v
\end{equation}
\end{lemma}

The matrix $\mH_{\{\tau, n, \gG\}}$ depends on the hyperparameter $\tau$, number of players $n$, and the graph $\gG$. When these variables are clear from the context, we drop them and write $v_{\tau}^{*} = \mH v$. Please refer to \cite{hn_value} for the proof of Lemma \ref{lemma:matrix_form}.


With the matrix form, we can define the limit game.
\begin{definition} \label{def:limit_game}
Given a game $(N, v, \gG)$, its limit game $(N, \tilde v, \gG)$ is defined by 
\begin{equation} \label{eq:limit_game}
    \tilde v = \lim_{p \to \infty} \mH^{p} v
\end{equation}
\end{definition}

Notice that although the matrix $\mH$ is constructed from the associated game and depends on $\tau$, the powers of $\mH$ actually converge to a limit independent from $\tau$, when $\tau$ is sufficiently small. The general condition depends on the actual graph, but $ 0 < \tau < \frac{2}{n}$ is proven to be sufficient for the complete graph case \cite{associated_shapley}. As we discussed in Section \ref{subsec:hn_value}, the limit game can be seen as constructing associated games repeatedly until the characteristic function converges. 

An axiomatic characterization of the HN value regarding its uniqueness is given by the following theorem based on the limit game. The associated consistency is the core property related to this work. We encourage the readers to check \cite{hn_value} for the other two properties.

\begin{theorem} \label{thm:main}
There exists a unique solution $\phi$ that verifies the associated consistency, i.e. $\phi_i(N, v, \gG) = \phi_i(N, v^*_{\tau}, \gG)$, inessential game, and continuity. $\phi$ is given by
\begin{equation}
    \phi_i(N, v, \gG) = \tilde v (\{ i\})
\end{equation}
\end{theorem}

\subsection{The algorithm for computing the HN value} \label{app:hn_compute}

We show the algorithm for 
\computehnmc (Algorithm \ref{alg:computehnmc}) mentioned in Section \ref{subsec:hn_explain}. The algorithm is a combination of Equation \ref{eq:matrix_form}, \ref{eq:limit_game}, and \ref{eq:feature_score}. 



\begin{algorithm}[htb]
  \caption{The \computehnmc Function}
  \label{alg:computehnmc}
\begin{algorithmic}
  \STATE {\bfseries Input:} Graph instance $\gG$ with nodes $\V = \{u_1, \dots, u_n \}$, characteristic function $v$, hyperparameter $\tau$, maximum sample size $m$, number of samples $J$
\STATE Let $\psi_1, \dots, \psi_n$ be $n$ empty lists
\FOR{$j=1$ {\bfseries to} $J$}
\STATE Sample $g_{S^j}$ from $\gG$ s.t. $S^j = \{u_{j_{1}}, \dots, u_{j_{l}}\}$ and $l<m$ 
\STATE $\phi^j = \computehn(g_{S^j}, S^j, v(\cdot), \tau)$
\FOR {$k=1$ {\bfseries to} $l$}
\STATE Append $\phi^j_k$ to $\psi_{j_k}$
\ENDFOR
\ENDFOR
\STATE Set $\phi_i$ to be the mean of $\psi_i$
\STATE \textbf{Return:} $\phi$
\end{algorithmic}
\end{algorithm}



\section{The Myerson value and the C-Shapley value} \label{app:myerson_and_c_shapley}
\subsection{The Myerson value} \label{app:myerson}
In the study of cooperative games, \cite{myerson} proposed to characterize the cooperation possibilities between players using a graph structure $\gG$, which leads to the communication structure introduced in Section \ref{subsec:cooperative_game} and the Myerson value as a solution for this special type of games $(N, v, \gG)$. The Myerson value is closely related to the Shapley value. In fact, it is the Shapley value on a transformed game where players are partitioned by the graph. We now formally introduce the partition and the transformed game. 

\begin{definition} [\textbf{Partition}]
Given a set of players $N$ and a graph $\gG$. For any coalition $S \subseteq N$, the partition of $S$ is denoted by $S / \gG$ and defined by 
\begin{equation*}
S / \gG = \{ \{ i \vert i \text{ and } j \text{ are connected in S by } \gG \} \vert j \in S\}
\end{equation*}
and a member of the set $S / \gG$ is called a component of $S$.
\end{definition}

\begin{definition} [\textbf{Transformed Game}]
Given a game $(N, v, \gG)$, we can transform it to a new game $v / \gG$ such that for all $S \subseteq N$
\begin{equation*}
(v / \gG)(S) = \sum_{T \in S / \gG} v(T)
\end{equation*}
\end{definition}

Intuitively, given a coalition $S$, the transformed game treats each connected component of $S$ as independent, evaluates them separately, and sums their payoff as the payoff of $S$.

The Shapley value has an axiomatic characterization that uniquely determines it as we introduced in Appendix \ref{app:shapley}. Likewise, the Myerson value was proposed to be a unique solution that satisfies the \textit{component efficiency} and the \textit{fairness} property defined below. 

\begin{property} [\textbf{Component Efficiency}] 
For a game $(N, v, \gG)$ and any connected component $S \in N / \gG$, a solution is component efficient if
\begin{equation*}
\sum_{i \in S} \phi_i(N, v, \gG)  = v(S) 
\end{equation*}
\end{property}

\begin{property} [\textbf{Fairness}]
For a game $(N, v, \gG)$ and any edge $(i, j)$ in $\gG$, let $\tilde \gG$ be $\gG$ with the edge $(i, j)$ removed, a solution is fair if
\begin{equation*}
\phi_i(N, v, \gG) - \phi_i(N, v, \tilde \gG) = \phi_j(N, v, \gG) - \phi_j(N, v, \tilde \gG)
\end{equation*}
\end{property}

The component efficiency property is an extension of the regular efficiency property to games with a communication structure. It requires efficiency to hold for each disconnected piece because these pieces are assumed as independent from each other. The fairness property states that if breaking an edge $(i, j)$ changes the value of player $i$, then the value of player $j$ should be changed by the same amount.

\begin{theorem}[\textbf{Myerson Value}]
There exists a unique solution $\phi$ of game $(N, v, \gG)$ satisfying component efficiency and fairness. With $\tilde \phi$ represents the Shapley value, the solution is given by the formula
\begin{equation*}
\phi(N, v, \gG) = \tilde \phi(N, (v / \gG))
\end{equation*}
\end{theorem}

For games with a communication structure, the Myerson value is a better choice than the Shapley value as it uses the graph structure. However, it also suffers from some criticisms. For example, the fairness assumption may not be realistic. When an existing edge is broken, the value changes for players on the two edge ends can be asymmetric. Intuitively, if the edge connects a popular hub player $i$ to a leaf player $j$, then the change of $i$ can be less significant than $j$ since $j$ becomes isolated when $(i, j)$ is removed. This is also the case when the game value is used for model explanation. For example in Figure \ref{figure:motivation_examples} (b), when the edge ("good", "quite") is broken, the value of "quite" should change a lot. It used to contribute positively together with "good", and thus gets some payoff allocation, but it now becomes an isolated node, which is neutral by itself. On the other hand, the word "good" can still contribute positively by itself and interact with other nodes through its other edges, and thus its value shouldn't change too much. Because of such criticisms, we choose to use the HN value as our scoring function, which characterizes the value by associated consistency rather than fairness.

\subsection{The C-Shapley value} \label{app:c_shapley}
The Myerson value was also mentioned in \cite{l_and_c} for the model explanation on text, where the C-Shapley value was proposed as an approximation of the Shapley value, and it was claimed to be equal to the Myerson value. We have discussed why Shapley value and Myerson are not-ideal choices for explaining graph data in Section \ref{sec:motivation} and Appendix \ref{app:myerson}. These are partially the reason why our HN-value-based method is better than the C-Shapley value. However, the major reason why we don't do a direct comparison to the C-Shapley value as a baseline is that its formula only works for line graphs like sequence data, and not even all nodes in line graphs. In contrast, our target task is general graph prediction for graphs with possibly complicated topological structures.

We now clarify a mistake of the C-Shapley value formula and explain why it won't work for general graphs. The notations are following the \cite{l_and_c}, where $d$ is the number of players corresponding to $n$ in our notation, and $[d]$ corresponding to $N$.

The formula for the C-Shapley value is given in Equation 6 in Definition 2 in the paper, and it is stated for "a graph G" without mentioning any assumptions of the graph. However, from the proof of this formula in Appendix B.2 in the paper, the line graph assumption can be seen in two places. The first place is Equation 20, where the set $\mathcal{C}$ is explicitly defined only for subsequences. The second place is Equation 22, the first line converts $\sum_{A:U_S(A) = U}$ to $\sum_{i=0}^{d - |U| - 2}$, which is implicitly saying $V_S(A)$ can be picked from all $d$ but $|U| - 2$ nodes. However, this conversion is only possible when there are exactly 2 edges between $U$ and $[d] \backslash U$, i.e. the middle part of a line graph. If there are $l$ edges between $U$ and $[d] \backslash U$, then the summation should go up to $d - |U| - l$. When $l = 0$, i.e. $U$ equals $[d]$ or a connected component of $[d]$, no partition is needed and the coefficient simply evaluates to 1. By correcting all these cases, the final formula for the C-Shapley value coefficients of marginal contributions thus becomes


\begin{align}
    &\sum_{i=0}^{d - |U| - l} \frac{1}{\binom{d - 1}{i + |U| - 1}} \binom{d - |U| - l}{i} \\
    &= \frac{d}{(|U| + l)\binom{|U| + l - 1}{|U| - 1} } \\
    &= \frac{dl}{(|U| + l)(|U| + l - 1) \cdots |U|}
\end{align}
for $l > 0$, and 1 for $l = 0$.

The correct formula for the C-Shapley value of general graphs will be
\begin{align} \label{eq:corrected_c_shapley}
   \phi_{X}(i) = 
   \begin{cases}
    \sum_{U \in \mathcal{C}}\frac{l}{(|U| + l) \cdots |U|} m_{X}(U, i) & \text{if } l > 0  \nonumber \\
    \frac{1}{d} & \text{if } l = 0
    \end{cases} \\
\end{align}
with $l$ represents the edges between $U$ and $[d] \backslash U$ and $\mathcal{C}$ represents all connected subgraphs in $[d]$ containing $i$.

\begin{figure*}[t]
\begin{center}
\includegraphics[width=0.98\textwidth]{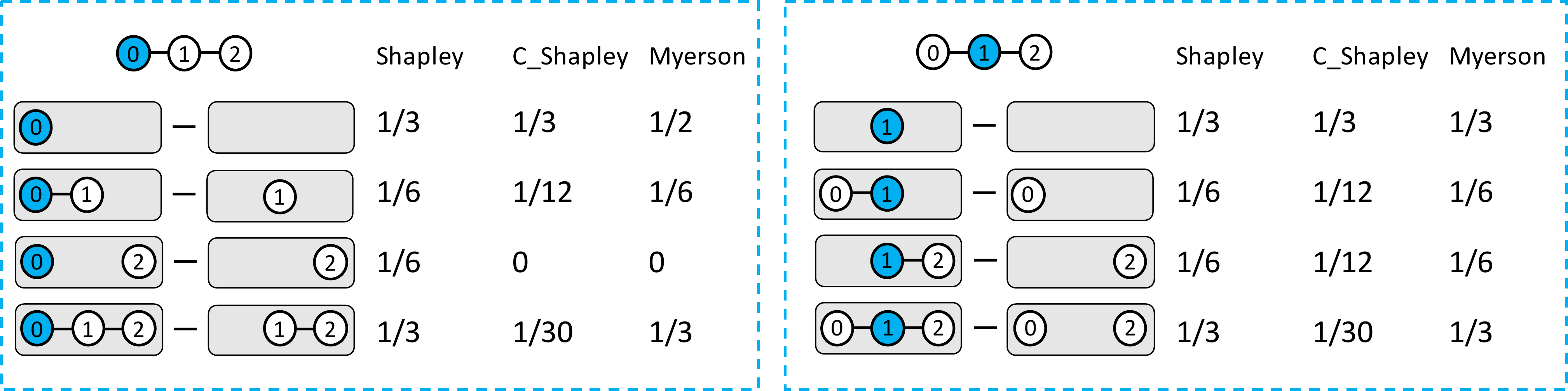}
\end{center}
\caption{A toy 3-node graph example for comparing the mariginal contribution coefficients between the Shapley, the C-Shapley, and the Myerson value. \textbf{(a)} Value computation for node $0$ (left).  \textbf{(b)} Value computation for node $1$ (right).}
\label{figure:c_shapley}
\end{figure*}

To verify this formula with the 3-node toy graph in Figure \ref{figure:c_shapley}. When computing the value of node 0 (left), the three connected components containing 0 are $\mathcal{C} = \{ \{0\}, \{0, 1\}, \{0, 1, 2\} \}$. Since 0 is an end node and has no leaf nodes to its left, $l$ for these three components  will be 1, 1, and 0 respectively. According to our new formula in Equation \ref{eq:corrected_c_shapley}, the coefficients will be $\frac{1}{2}$, $\frac{1}{6}$, and $\frac{1}{3}$ respectively, with the disconnected $\{0, 2\}$ case removed. This matches the original idea of Myerson value, where the $\{0, 2\} - \{2\}$ case is reduced to the $\{0\} - \emptyset$ case, which turns the Shapley coefficients from $[\frac{1}{3}, \frac{1}{6}, \frac{1}{6}, \frac{1}{3}]$ to $[\frac{1}{3} + \frac{1}{6}, \frac{1}{6}, \frac{1}{6} - \frac{1}{6}, \frac{1}{3}]$, which is $[\frac{1}{2}, \frac{1}{6}, 0, \frac{1}{3}]$. However, the original C-Shapley formula from Equation 6 in the \cite{l_and_c} evaluates to $[\frac{1}{3}, \frac{1}{12}, 0, \frac{1}{30}]$, which doesn't match the Myerson value and not even sum up to 1. Another example of computing the value of node 1 is shown in Figure \ref{figure:c_shapley} right.

The C-Shapley, even with the correct formula, eventually boils down to an approximation of the Shapley value or the Myerson value, which as we discussed are less ideal than the HN value. Also, the correct formula in Equation \ref{eq:corrected_c_shapley} requires generating all possible subgraphs $U$ containing the node $i$ and specify the edges between $U$ and $[d] \backslash U$. This makes the computation very complicated, we thus skip the comparison to the C-Shapley value.

\section{Use the graph structure via an L-hop cutoff} \label{app:cutoff}
Although the Shapley value itself is not structure-aware, we do note the existing Shapley-value-based GNN explanation methods use an L-hop cutoff to help approximate the Shapley value \cite{subgraphx, graphsvx}. Technically, this operation uses the graph structure, so we can't strictly refer to these explanation methods as not structure-aware. However, we argue that the L-hop cutoff is a naive way of utilizing the graph structure. It has several concerns, and it is not the same structure-aware as the HN value.

The L-hop cutoff approximates the Shapley value of node $i$ by considering only the L-hop neighbors of $i$ when explaining an L-layer GNN. The rationale of this operation is that an L-layer GNNs only propagate messages within L-hops so a node more than L-hop away from $i$ has never passed any messages to $i$ which means no interactions are possible. In existing Shapley-value-based GNN explanation methods, this L-hop cutoff operation was meant for reducing the exponentially growing computations of the Shapley value, and the ultimate goal is still to compute the Shapley value. The L-hop cutoff operation has several issues making it a less desirable choice. \textbf{1)} Even meant to save computation, there are still many nodes involved in the computation after applying the L-hop cutoff since the number of nodes grows exponentially as L grows. For advanced GNNs, the L can be large. When L is larger than the diameter of the graph, which is actually the case for many recent deep GNNs, the L-hop cutoff is not effective anymore. \textbf{2)} When constructing coalitions of nodes within the local graph of L-hops, the computation still follows the Shapley value formula. This means the useful graph structure information among these nodes is forfeited which causes the structure-awareness concern of Shapley value as we discussed in Section \ref{sec:motivation},

\section{More explanation visualizations} \label{app:visualization}
\begin{figure*}[ht]
\begin{center}
\includegraphics[width=\textwidth]{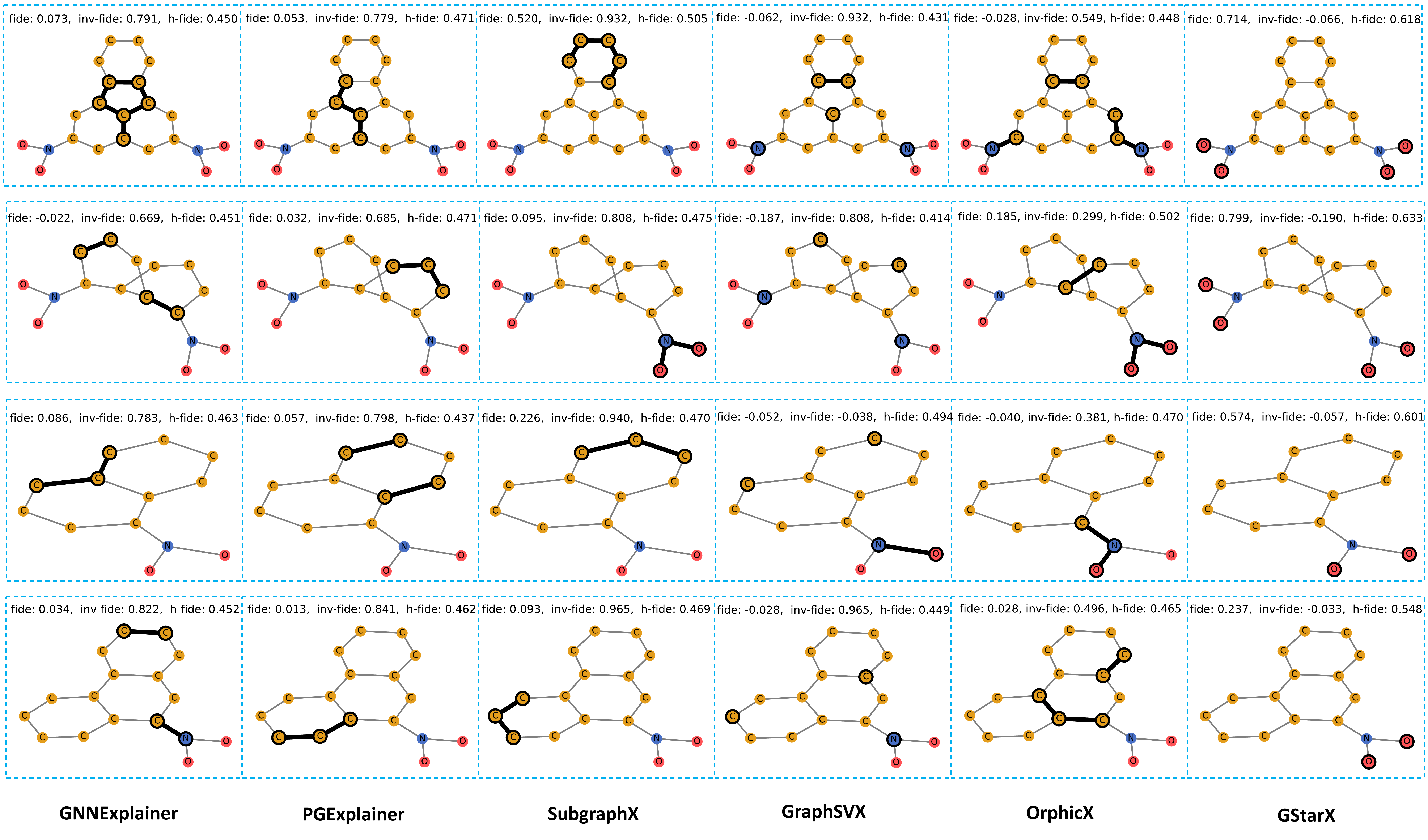}
\end{center}
\caption{Explanations on a mutagenic molecule from the \mutag{} dataset. Carbon atoms (C) are in yellow, nitrogen atoms (N) are in blue, and oxygen atoms (O) are in red. We use dark outlines to indicate the selected subgraph explanation and report the \fidelity (fide), \invfidelity (inv-fide), and \harfidelity (h-fide) of each explanation. \method{} gives a significant better explanation than other methods in terms of these metrics.}
\label{figure:more_examples}
\end{figure*}

Under the same setting as Figure \ref{figure:mutag_examples}, we visualize more explanations in \ref{figure:more_examples}.


\begin{figure*}[ht] 
\begin{center}
\includegraphics[width=\textwidth]{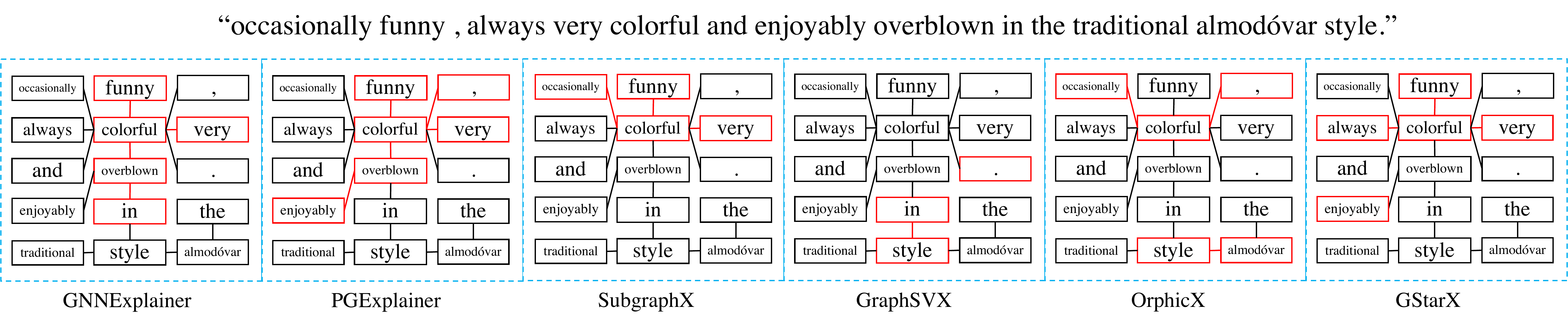}
\end{center}
\caption{Explanations on sentences from \sst{}. The sentence is predicted to be positive sentiment. Red outlines indicate the selected nodes/edges as the explanation. \method{} identifies the sentiment words more accurately compared to baselines.}
\label{figure:sst_more_examples}
\end{figure*}

\end{document}